# AN EXPERT SYSTEM FOR AUTOMATIC READING OF A TEXT WRITTEN IN STANDARD ARABIC

Tebbi Hanane[1] and Azzoune Hamid[2]

[1]LRIA, Option: Representation of Knowledge and systems of Inference, USTHB, Algiers
[2] LRIA, Option: Representation of Knowledge and systems of Inference, USTHB, Algiers

## ABSTRACT

*In this work we present our expert system of Automatic reading or speech synthesis based on a text written in Standard Arabic, our work is carried out in two great stages: the creation of the sound data base, and the transformation of the written text into speech (Text To Speech TTS). This transformation is done firstly by a Phonetic Orthographical Transcription (POT) of any written Standard Arabic text with the aim of transforming it into his corresponding phonetics sequence, and secondly by the generation of the voice signal which corresponds to the chain transcribed. We spread out the different of conception of the system, as well as the results obtained compared to others works studied to realize TTS based on Standard Arabic.*

## 1. INTRODUCTION

The automatic speech generation is a complex task because of the variability intra and interlocutor of the voice signal. In computer science the difficulty of modeling the speech signal is due to the fact that we don't know till now how to model very well the enormous mass of knowledge and information useful for the speech synthesis. So we have made a choice to use an expert system to model that knowledge to build a robust system which can really read a text written in a language chosen especially in Standard Arabic. To obtain a better organization of our work, we defined our direct aims. We divided our modeling into three essential stages; the signal analysis, the phonetic orthographical transcription (POT) and finally the synthesis of the textual representation written in Standard Arabic. Therefore the finality considered here is that the user can understand the different sentences transcribed and synthesized which will be pronounced with a clear and high quality way.

## 2. OVERVIEW OF SPEECH SYNTHESIS SYSTEMS

At the present time, we can judge that works carried out in the same context of ours are still not really colossal, and this is because of the complexity of the language itself. And if this some works exists, they are based on the same principle as transcribers of the other languages (French, English, etc.) [1], nevertheless, the efforts undertaken are encouraging and open a large window to follow research task in this field.

– TTS system MBROLA [2] which use the code SAMPA during the stage of transcription, in this case the user     must respect the form of the SAMPA code which is not a universal code;
– Work of S. BALOUL [3] who represents a concrete example of transcription of words; based on morphological analysis, and on the studies of pauses to generate the pronunciation of the texts.

– SYAMSA (SYstème d'Analyse Morphosyntaxique de l'Arabe), realized by SAROH [4]. According to him, "the phonetisation of the Arabic language is based in particularly on the use of lexicons and on a morphological analyzer for the generation of the different forms of a word. In addition, they are the phenomena of interaction among the words (connection, elision, etc.) and the phenomena of assimilation which suggest the uses of phonological rules "[5]. This tool ensures for each word in entry, the root which correspond to it as well as the morphological and phonetic representations of the root.
– The GHAZALI [6] project which was carried out within the IRSIT (Institut Régional des Sciences Informatiques et des Télécommunications) of Tunisia, it is based on a transcription work which fits inside the framework of the realization of a TTS system. The characteristic of this system is shown in the use of a set of rules in the emphasis propagation.
– SYNTHAR+ [7] which is a tool studied by Z. ZEMIRLI within the NII (the National Institute of Informatics of Algiers), it ensures the transcription step for a TTS system so that it transmits the phonetic representation to the MULTIVOX synthesizer. It should be known that SYNTHAR+ is based on a morphological analysis before realizing the transcription.

## 3. DESIGN OF THE SYSTEM

The uses of the concept of code and the introduction of high levels in analysis (morphological, syntactic, and pragmatic.) makes the transcription task so difficult and requires deepened studies of the language itself. The difference in our work compared to all that exists is in the transcription using graphemes, i.e. the uses of the Arabic characters as basic units to transcribe directly the text, indeed modeled by an expert system.

Thus the role of an expert system is then to infer generally rules like if/then. To build this system our work targets the following steps:

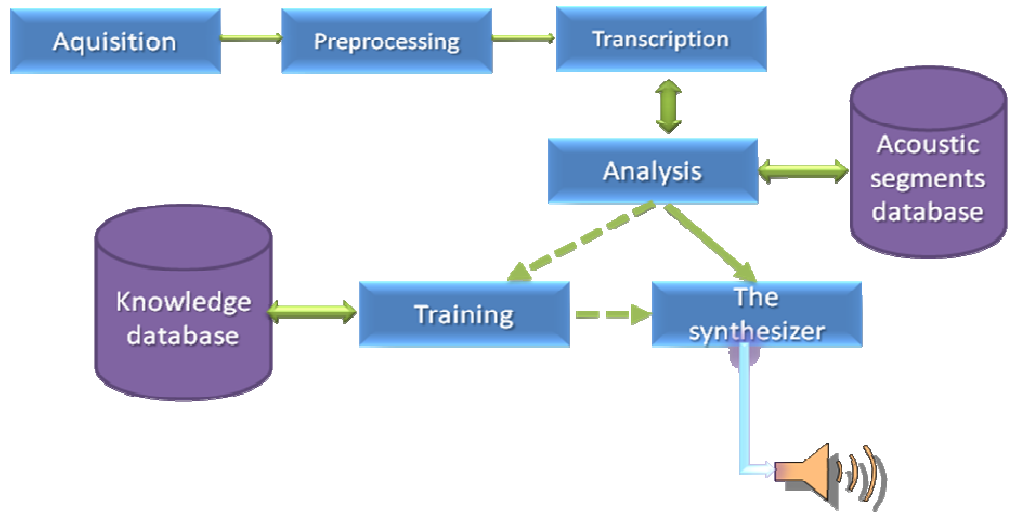

Figure 1. General schema of our system of speech synthesis

### 3.1. Acquisition

It is the first step of our process of voice synthesis; it plays the role of the interface intermediate between the user and the system. During this time the user have to enter his set of phrases to be pronounced by the system. After that the automatic process can begin.

### 3.2. Pre-processing and standardization of the text

In this phase any form of text is transformed to it literal, perfectly disambiguated form [1]. In particular, this module draft problems of formats, dates and hours, abbreviations, numbers, currencies, addresses, emails…etc. It was shown that majority errors of conversion grapheme/phoneme, for the best operational systems came from the proper names and the exceptions that they pose [8]. Here are some examples of the processing carried out in our case:

– The replacement of each composed character by its equivalents
     Example:  ﻵ ⟶ ﺍ ﻝ
– Consultation of the exceptions lexicon to eliminate the special words.
– The application of the transcription rules established for the language. This module must be able to process grammatical complexes tasks to identify the rules of transcription which will be used in a considered context.

### 3.3. Transcription

Two approaches are generally used here:

– **The use of the lexicon:**

   In this case we must assign to each word in input, the pronunciation which corresponds to it without taking in account its context. The speed, flexibility and simplicity are the main advantages of this approach.
   It has been shown that the majority errors of conversion grapheme/phoneme, for the best operating systems came from the proper names and the exceptions that they pose [9]. Exceptions are words that are not read based on well determined rules of writing; because of this they represent the famous example of this approach.

Table 1: a few words of exceptions [5]

| Les mots d'exception | Prononciation correcte | Transcription en API |
|---|---|---|
| هذا | هاذَا | haaมa |
| ذلك | ذَالِك | มaalika |
| كذلك | كذَالِك | kaมaalika |
| يٰأيها | يٰأَيُّها | yaaeayuhaa |
| هؤلاء | هٰؤُلاء | haaulaaei |
| يــــــن | يَاسِين | yaasiin° |

- **The use of rules**

  In this approach each grapheme is converted to a phoneme depending on the context using a set of rewriting rules. The main advantage of this approach come out in the ability to model the linguistic knowledge of human beings by a set of rules that can be incorporated in expert systems. Each of these rules has the following form:

  [Phoneme] = {LG (LeftContext) + {C (character)} + {RD (Right Context)}

  Here is a concrete example of transcription rules [11]:
  | # | Is a sign of beginning of sentence, |
  |---|---|
  | $ | Is a sign of end of sentence, |
  | § | Is one end of word, |
  | C | Is a consonant, |
  | V | Is a vowel, |
  | SC | Is a Solar Consonant and LC is a Lunar Consonant |

  [UU] = {CS} + {و} + { ' } [uu]= §+ {} + { و ' }
  [UU] = {CL} + {و} + { ' } [uu]= $ {و } + { ' }

  - When the -" و " is preceded by the vowel/ / ' and followed by a consonant, we obtain the long vowel [uu].
  - When the " و " is preceded by the vowel / / ' in final position, we obtain the long vowel [uu].

The second type represents the diphones which are extracted from the set of artificial words and which do not have a meaning, also called logatomes, or words porters. Each of these is used to extract a single diphone i.e. each logatome contains one and only one diphone and this in order to ensure a degree of independence between this diphone and its context. Here are a few examples of logatomes used :
  C represents a consonant
  V a vowel
  # A silence for the beginning or end of word

| Logatomes | Diphones | Exemples |
|---|---|---|
| #katatv# | [v#] | #katat **fatha**_# |
| | | #katat **dama**_# |
| | | #katat **kasra**_# |
| | | #katat **madd fatha**_# |
| | | #katat **madd dama**_# |
| | | #katat **madd kasra**_# |
| | | |
| # taccata # | [cc] | # ta **b_b** ata # |
| #cata# | [#c] | #_ **t** ata# |
| #katac# | [c#] | #kata **f**_# |
| #acvta# | [cv] | #a **b_dama** ta# |
| #atvca# | [vc] | #at **kasra_n** a# |

## 3.4. Analysis (the extraction of the characteristics)

The aim of the analysis of the voice signal is to extracts the acoustic vectors which will be used in the stage of synthesis follows. In this step the voice signal is transformed into a sequence of acoustic vectors on the way to decrease redundancy and the amount of data to be processed. And then a spectral analysis by the discrete Fourier transform is performed on a signal frame (typically of size 20 or 30 ms) [9]. In this frame, the voice signal is considered to be sufficiently stable and we extract a vector of parameters considered to be enough for the good operation of the voice signal. In the speech synthesis, the characteristics extraction step, commonly known as the step of analysis, can be achieved in several ways. Indeed, the acoustic vectors are usually extracted using methods such as temporal encoding predictive linear (Linear Predictive Coding LPC) or Cepstrales methods as the MFCC encoding (Mel Frequency Cepstral Coding), as well as the process of segmentation, etc. This process delimits on the acoustic signal a set of segments characterized by labels belonging to the phonetics alphabet of the language under consideration.

At present, the segmentation completely automatic of a voice signal remains a fastidious task. Indeed, looking at the complexity acoustico-phonetics phenomena being studied, this activity requires often a manual intervention. Generally, the methods that perform the segmentation of the acoustic wave are divided into two great classes:

- The first include all the methods which allow segmenting a voice signal without a priori knowledge of the linguistics content of this signal. These methods split the voice signal into a set of zones homogeneous and stable;
- The second class includes all the methods which allow segmenting the voice signal basing on a priori linguistic description (typically on phonemes) of this signal. These methods of segmentation are revealed like methods with linguistically constraints. In our state, we have opted for the second technique using a tool of voice signal analysis which is PRAAT [10] so to slice manually the speech signal in a succession of segments, each one associated with an element acoustic unit (phoneme or diaphone).

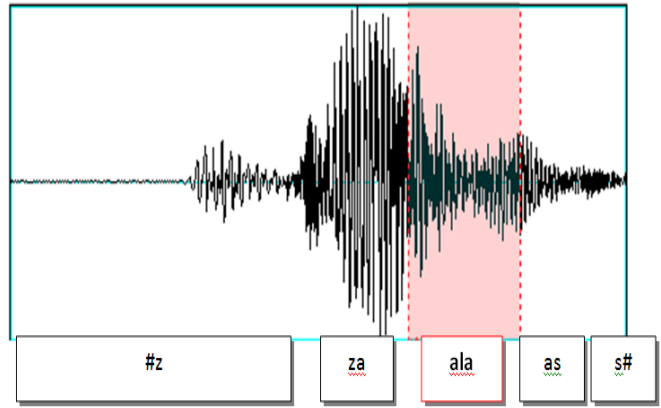

Figure 2.  Decomposition in polysound of the word [ جلس ]

### 3.5. Creation of the sound database

Majority of work curried out in the field of the spoken communication required often the recording, and the handling of corpuses of continuous speech, and that to curry out studies on the contextual effects, one the phonetic indices, and variability  intra and inter-speaker. Our corpus is modeled by the diagram of class1 as follow:

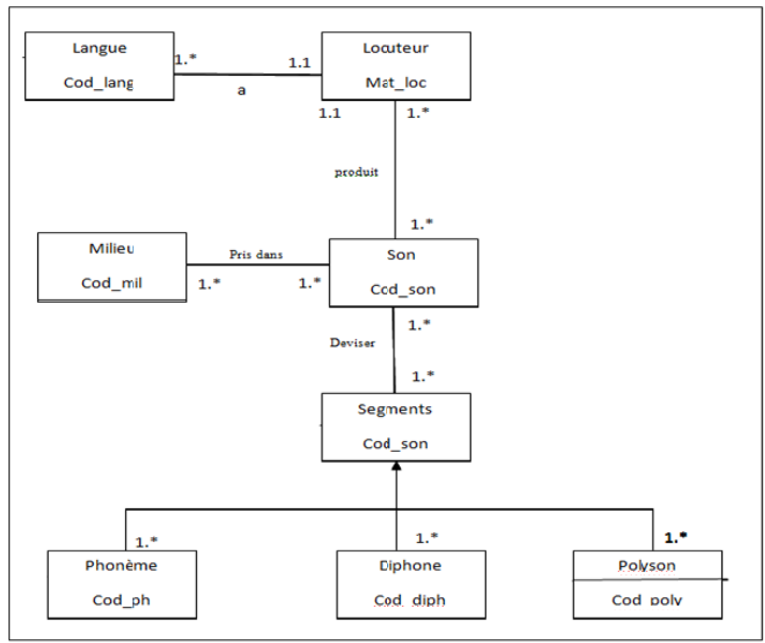

Figure 3. The diagram of class of our sound database

### 3.6. Voice generation

This step commonly known as speech synthesis consist of a transformation of the phonetic sequence (which represents the pronunciation of the written text) resulting from the

---

1: It is a diagram used in UML(Unified Modeling Language) which is used in object oriented modeling transcription step to its substance, i.e. to it acoustics realization. During this step we choose from our sound database the units (phonemes, diphones) most suitable to build the sentence to generate, this means that we will create a automatic function of reading, so that in the end, the user has only to listen to the synthetic sentences. To improve the quality of the generated synthesis, we increase each time the sound unit only. The general silhouette of our expert system can be as follow:

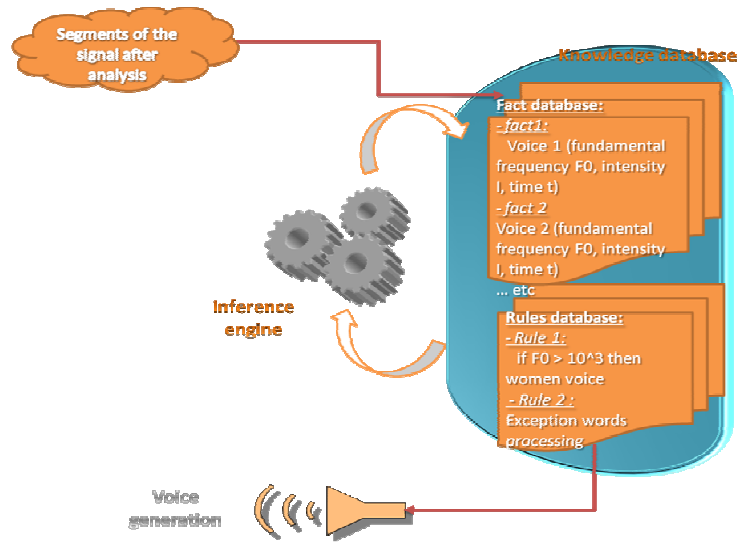

Figure 4. General diagram of our expert system

In our case we have used the method of synthesis by concatenation and here is our reading function:

```
position= find (grapheme[ ig],API ) ;
if((grapheme[ig]=='أ')&& (grapheme[ig+1]=='ل'))
    {
        mp2->FileName="C:\\son_hanane\\alif.wav";
        mp2->Open();
        mp2->Wait=true;
        mp2->Play();
    ig=ig+2;
        position=find (grapheme[ig],API);
        if(API[position][1]=='L')
        {
            mp2->FileName="C:\\son_hanane\\l.wav";
            mp2->Open();
            mp2->Wait=true;
            mp2->Play();
            mp2->FileName=API[position][2];
            mp2->Open();
            mp2->Wait=true;

            mp2->Play();
        ig++;
        }
```

```
else
    {
      if(API[position][1]=='S')
      {
      mp2->FileName=API[position][2];
      mp2->Open();
      mp2->Wait=true;
      mp2->Play();
      ig++;
      }
    }
  }
```

## 4. TESTS AND RESULTS

To test the performances of our TTS system based on Standard Arabic language, we have chosen a set of sentences which we judged like reference since they contain almost the different possible combinations specific to the language itself. To calculate the success rate (SR) associated with each sentence tested; we got the following formula:

$$SR = \frac{Number\,of\,phrase\,well\,prononced}{Number\,of\,phrases\,tested} * 100\%$$

The system present in general a SR of 96 % for the set of the sentences tested. Results obtained are summarized by the following table:

Table 2. Rate of success for a sample of selected sentences

| Majority content | POT | Synthesis by Phonemes | Synthesis by Diphones |
|---|---|---|---|
| Short vowels | 100% | 95% | / |
| Long vowels | 100% | 95% | / |
| Solar consonants | 100% | 97% | / |
| Lunar consonants | 100% | 95% | / |
| Isolated words | 100% | 80% | 90% |
| Sentences | 100% | 75% | 85% |
| Numbers | 90% | 95% | 100% |
| Exception Words | 100% | / | / |

## 5. CONCLUSION

Synthesis systems from text written in Standard Arabic represent a research area very active but difficult to implement. In order to facilitate this task we have proposed a tool, an expert system for voice generation based on Standard Arabic text has been developed, this system uses a set of sub phonetic elements as the synthesis units to allow synthesis of limited-vocabulary speech of good quality. The synthesis units have been defined after a careful study of the phonetic properties of modern Standard Arabic, and they consist of central steady-state portions of vowels, central steady-state portions of consonants, vowel-consonant and consonant-vowel transitions, and some allophones. A text-to-speech system which uses this method has also been explored. The input of the system is usual Arabic spelling with diacritics and/or simple numeric expressions.

Synthesis is controlled by several text-to-speech rules within the rule database of the expert system, which are formulated and developed as algorithms more suited for computer handling of the synthesis process. The rules are required for converting the input text into phonemes, converting the phonemes into phonetic units, generating the synthesis units from the phonetic units, and concatenating the synthesis units to form spoken messages. The suitability of the method for synthesizing Arabic has been shown by realizing all its functions on a personal computer and by conducting understandability test on synthesized speech. So, we have detailed the different components which represents the basic blocks of our TTS system based on a written text in Standard Arabic and our modeling of it with the use of an expert system, this tool which is a  fruits of the artificial intelligence remain less used in the field of  automatic speech processing. This fact encouraged us to explore this world trying to give a little push to the research done in this multidisciplinary field.